\documentclass[11pt,a4paper]{article}

\usepackage[utf8]{inputenc}
\usepackage[T1]{fontenc}
\usepackage[left=2.5cm, right=2.5cm, top=2.5cm, bottom=2.5cm]{geometry}
\usepackage{amsmath, amssymb, amsthm}
\usepackage{graphicx}
\usepackage{booktabs}
\usepackage{tabularx}
\usepackage[justification=raggedright, singlelinecheck=false, skip=0pt]{caption}
\usepackage{titlesec}
\usepackage{enumitem}
\usepackage{xcolor}
\usepackage{authblk}
\usepackage{lineno}
\usepackage{float}
\usepackage{fancyhdr}
\usepackage{hyperref}

\renewenvironment{abstract}
  {\noindent\small\textbf{Abstract:}\enspace}
  {\par\vskip 1em}

\newcommand{\keywords}[1]{
  \noindent\begin{flushleft}\textbf{Keywords:}\enspace #1\end{flushleft}
  \nointerlineskip
  \noindent\rule{\textwidth}{0.8pt}
  \vskip 2em
}

\makeatletter
\renewcommand{\maketitle}{
  \begin{flushleft}
    \nointerlineskip
    \noindent\rule{\textwidth}{0.8pt}
    {\huge \bfseries \@title \par}
    \vskip 2em
    {\@author \par}
    \vskip 1.5em
  \end{flushleft}
}

\makeatother

\theoremstyle{plain}

\theoremstyle{definition}

\titleformat{\section}{\normalfont\large\bfseries}{\thesection.}{0.5em}{}
\titlespacing{\section}{0pt}{3.5ex plus 1ex minus .2ex}{0.5ex}

\titleformat{\subsection}{\normalfont\normalsize\itshape}{\thesubsection.}{0.5em}{}
\titlespacing{\subsection}{0pt}{2ex plus 0.5ex minus .2ex}{0.1ex}

\titleformat{\subsubsection}{\normalfont\normalsize}{\thesubsubsection.}{0.5em}{}
\titlespacing{\subsubsection}{0pt}{0.5ex plus 0.2ex minus .1ex}{0.1ex}

\title{Classifier-Guided Latent Diffusion for Small-Molecule Generation}
\author[1,2,3]{\textbf{Ali Forooghi, Luis Rueda, Alioune Ngom}}
\affil[1,2,3]{School of Computer Science; University of Windsor, Windsor, Ontario, Canada}
\date{}

\begin{document}

\maketitle
\thispagestyle{fancy}

\begin{abstract}
Diffusion models have emerged as a promising class of generative models for molecular design because of their ability to learn complex chemical distributions and support conditional or guided generation. In this work, we investigate a latent diffusion pipeline for property-guided small-molecule generation. The proposed framework represents canonicalized SMILES strings using a frozen ChemBERTa encoder, trains a Transformer-based decoder to reconstruct SELFIES strings from molecular latent embeddings, and learns an unconditional denoising diffusion model in the ChemBERTa latent space. To enable property-directed generation, a differentiable latent property predictor is used to guide the diffusion sampling process. Generated SELFIES strings are converted to SMILES, validated using RDKit, and evaluated using molecular validity, uniqueness, novelty, and predicted property scores. We study guidance toward two drug-relevant binary endpoints: blood--brain barrier permeability and inhibition of human $\beta$-secretase 1. The results suggest that latent diffusion over pretrained molecular embeddings can generate chemically valid and novel molecules while enabling controllable steering toward desired molecular properties.
\end{abstract}


\section{Introduction}

Machine-learning models for molecular generation have become increasingly important in drug discovery and materials design, where the chemical search space is too large for exhaustive experimental screening. Generative models can support inverse molecular design by proposing candidate molecules under structural, physicochemical, or biological constraints \cite{benjamin2018inversesurvey,Du2024MachineLG}. Among recent generative approaches, diffusion models have attracted growing attention because they combine expressive distribution modeling with flexible mechanisms for conditional and guided generation. Recent surveys on diffusion models for de novo molecular design show that these methods have been applied across multiple molecular representations, including strings, graphs, three-dimensional structures, and latent spaces, with applications ranging from unconditional molecule generation to property-guided optimization and structure-aware design \cite{wang2025diffmolsurvey,Qin20253dsurvey,alakhdar2024diffsurvey}.

Diffusion models are particularly suitable for property-guided molecular generation because they generate samples through an iterative denoising process rather than a single-step mapping. In the denoising diffusion probabilistic model formulation, a neural network learns to reverse a gradual Gaussian noising process, making the framework naturally compatible with continuous molecular representations \cite{SohlDickstein2015DeepUL,Ho2020denoisingdiff}. In molecular design, this iterative sampling process is useful because the generation trajectory can be gradually steered toward desired properties. Predictor-based guidance further enables flexible property control by using gradients from a property predictor during sampling, without requiring a separately trained conditional generative model for each target property.

Existing molecular diffusion models differ substantially in the representation on which diffusion is performed. Three-dimensional diffusion models operate on atom types and spatial coordinates and are well suited to conformer generation and structure-based molecular design \cite{Vignac2023MiDiMG, Liu2025NExTMol3D,Feng2024UniGem}, while graph-based diffusion models generate molecular graphs through atom and bond transitions \cite{liu2024graphdit, Qiang2023CoarsetoFineAH, Lee2022ExploringCS}. In addition, some recent approaches use hybrid graph--geometry representations that combine molecular connectivity with three-dimensional structural information, allowing the model to capture both chemical topology and spatial molecular organization \cite{Hua2023MUDiffUD, huang2025midi, Peng2023MolDiffAT}. These approaches are chemically expressive, but they often require specialized architectures for geometric equivariance, permutation invariance, or discrete graph transitions. Latent diffusion provides a complementary alternative by performing the generative process in a learned continuous representation rather than directly in molecular graph, coordinate, or string space \cite{rombach2022highresimgae}. This can reduce the complexity of direct molecular generation while allowing the model to reuse pretrained molecular representations.

\begin{figure}[!t]
    \centering
    \includegraphics[
        width=\linewidth
    ]{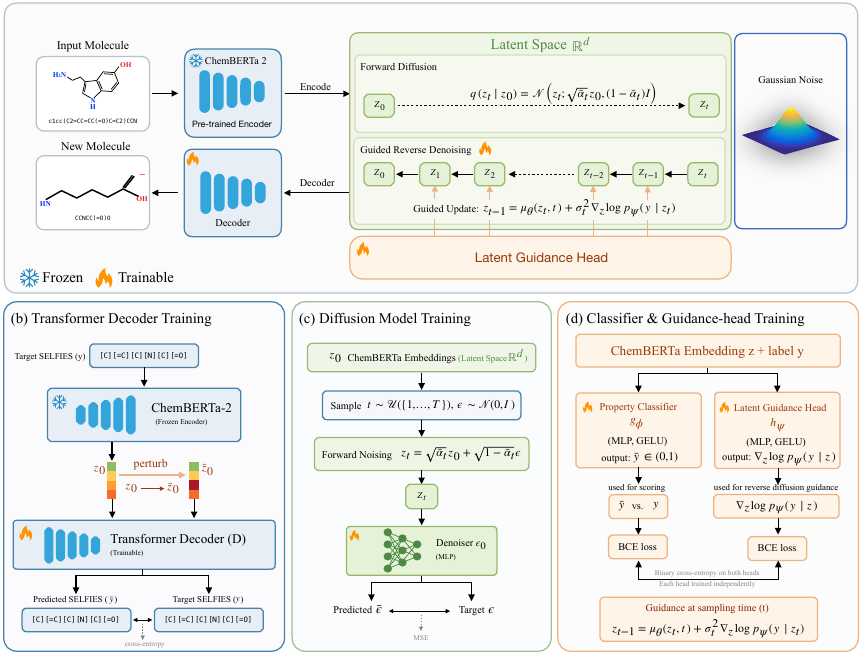}
    \caption{Property-guided molecular generation framework. (a) Molecules are encoded into the ChemBERTa-2 latent space, optimized by classifier-guided reverse diffusion, decoded to molecular strings, and validated with RDKit. (b) The transformer decoder reconstructs SELFIES from perturbed latent embeddings. (c) The diffusion model predicts noise added to latent representations. (d) The property classifier scores generated molecules, while the guidance head steers sampling toward target properties.}
    \label{fig:overview}
\end{figure}

In this work, we investigate a latent diffusion framework for property-guided molecular string generation. Instead of denoising atom coordinates or graph states directly, the proposed framework encodes canonicalized SMILES \cite{weininger1988smiles, weininger1989smiles}, strings using a frozen ChemBERTa encoder and learns an unconditional diffusion model over the resulting continuous molecular embeddings \cite{chithrananda2020chemberta, ahmad2022chemberta2}. Generated latent vectors are decoded into SELFIES \cite{krenn2020selfies} strings using a Transformer decoder and then converted to SMILES for evaluation. SELFIES are used as the decoder output representation because they provide a robust molecular string format that helps reduce syntactic invalidity during molecular reconstruction from latent space to sequence space.

The proposed framework is motivated by prior works on latent molecular diffusion and predictor-guided generation, where diffusion models are combined with property predictors to enable controllable small-molecule generation \cite{rombach2022highresimgae, Oestreich2024DrugDiffS}. While many latent diffusion approaches rely on learned variational autoencoder (VAE) representations, this work examines whether embeddings from a pretrained molecular language model can serve as an effective latent space for diffusion-based molecular generation. In this formulation, the diffusion model operates over continuous ChemBERTa embeddings, and a separate decoder maps generated latent representations back to molecular strings. In addition, whereas many previous molecular-generation frameworks have primarily emphasized optimization of physicochemical properties, such as molecular weight, logP, or related drug-likeness descriptors, this study evaluates guidance toward non-physicochemical, biologically relevant endpoints, namely blood--brain barrier permeability (BBBP) and inhibition of $\beta$-secretase 1 (BACE-1). Thus, the proposed framework explores both a distinct pretrained latent representation and property guidance toward biologically relevant molecular endpoints.

We evaluate the framework in both unconditional and guided generation settings. Generated molecules are assessed using validity, uniqueness, novelty, and predicted property scores. For guided generation, we study two drug-relevant binary endpoints from MoleculeNet benchmark \cite{wu2018moleculenet}: blood--brain barrier permeability and inhibition of human $\beta$-secretase 1 (BACE-1) \cite{subramanian2016bace, martins2012bbbp}. The main contributions of this work are: (i) a ChemBERTa-based latent diffusion pipeline for molecular string generation, (ii) a SELFIES decoder for reconstructing molecular strings from generated latent vectors, and (iii) an evaluation of predictor-guided latent sampling for steering generation toward selected molecular property endpoints. 

\section{Materials and Methods}

The proposed method performs molecular generation in the continuous latent space of a pretrained molecular language model. Canonicalized SMILES strings are first encoded into ChemBERTa embeddings, which serve as the training space for an unconditional denoising diffusion model. During sampling, latent vectors are generated through the reverse diffusion process and decoded into SELFIES strings, which are subsequently converted to SMILES for chemical validation. To enable property-guided generation, gradients from a pretrained property classifier and latent guidance head are incorporated into the reverse denoising process, steering generated molecules toward selected target properties while preserving the same pretrained latent representation and decoder. An overview of the proposed pipeline is shown in Figure~\ref{fig:overview}. The following subsections describe each component of the framework in more detail.

\begin{table}[!t]
\centering
\caption{Datasets used for generative and property-model training, including their sources and data-splitting strategies.}
\label{tab:dataset-sizes}
\scriptsize
\begin{tabularx}{\textwidth}{lllcX}
\toprule
\textbf{Dataset} &
\textbf{Source} &
\textbf{Role} &
\textbf{Molecules} &
\textbf{Split strategy} \\
\midrule
ZINC250K &
ZINC Dataset \cite{irwin2020zinc} &
Decoder and diffusion training &
249,455 &
Scaffold split train/validation/test split (80/10/10) \\

BBBP &
MoleculeNet \cite{wu2018moleculenet,martins2012bbbp} &
BBBP classifier and latent guidance head &
2,039 &
Scaffold split train/test split (80/20) \\

BACE-1 &
MoleculeNet \cite{wu2018moleculenet,subramanian2016bace} &
BACE-1 classifier and latent guidance head &
1,513 &
Scaffold split train/test split (80/20) \\
\bottomrule
\end{tabularx}
\end{table}
\subsection{Data and preprocessing}

The proposed framework uses two types of molecular datasets. ZINC250K \cite{irwin2020zinc} is used to train the molecular reconstruction decoder and latent diffusion model, while labeled molecular property datasets from the MoleculeNet benchmark \cite{wu2018moleculenet} are used to train the property-prediction and latent-guidance models. Property-guided generation is evaluated on two binary endpoints: blood--brain barrier permeability (BBBP) \cite{martins2012bbbp} and inhibition of human $\beta$-secretase 1 (BACE-1) \cite{subramanian2016bace}. Table~\ref{tab:dataset-sizes} summarizes the source, role, number of molecules, and data-splitting strategy used for each dataset.

All input molecules are represented as SMILES strings and standardized before model training. Invalid molecules are removed using RDKit\footnote{https://www.rdkit.org/}, and valid molecules are canonicalized to ensure a consistent molecular representation. Duplicate molecules are removed after canonicalization. For supervised property prediction, molecules are divided into training and validation sets using the splitting procedure described in Table~\ref{tab:dataset-sizes}. The data split was generated once using a fixed random seed and was retained for all corresponding model-training and validation experiments. Model training was not repeated using alternative data-splitting seeds.

For sequence reconstruction, canonical SMILES strings are converted to SELFIES representations. SELFIES are used as the decoder output format because they provide a robust molecular string representation that reduces syntactic invalidity during molecular generation. A vocabulary is constructed from the SELFIES tokens in the training set and augmented with special sequence tokens for padding, beginning-of-sequence, end-of-sequence, and unknown tokens.

\subsection{Molecular latent representation}

Each canonicalized SMILES string is encoded using a pretrained ChemBERTa model. The ChemBERTa encoder is kept frozen in order to preserve its pretrained molecular representation space and reduce the number of trainable parameters. Given a molecular string $x$, the encoder produces token-level hidden representations, which are aggregated using masked mean pooling to obtain a fixed-dimensional molecular embedding: 
\begin{equation}
z_0 = E(x) \in \mathbb{R}^{d},
\end{equation}
where $E$ denotes the frozen ChemBERTa encoder and $d$ is the embedding dimension. These continuous molecular embeddings serve as the latent space for decoder training, diffusion training, and property-guided generation.

\subsection{Embedding-to-SELFIES decoder}

To map latent molecular embeddings back to molecular strings, a Transformer-based decoder $D$ is trained to reconstruct SELFIES sequences from ChemBERTa embeddings. The decoder receives a molecular embedding as latent memory and autoregressively predicts the corresponding SELFIES token sequence. During training, teacher forcing is used to predict each SELFIES token from the previous ground-truth tokens and the latent representation.

To improve robustness to generated latent vectors, which may not lie exactly on the empirical ChemBERTa embedding manifold, the decoder is trained with both clean and perturbed latent embeddings. Given a clean latent embedding $z_0$, a perturbed latent embedding is defined as
\begin{equation}
\tilde{z}_0 = z_0 + \sigma \epsilon,
\qquad
\epsilon \sim \mathcal{N}(0,I),
\end{equation}
where $\sigma$ controls the magnitude of the latent perturbation. The decoder reconstruction objective is defined as
\begin{equation}
\mathcal{L}_{\mathrm{dec}}
=
\mathrm{CE}\left(D(z_0), y\right)
+
\lambda
\mathrm{CE}\left(D(\tilde{z}_0), y\right),
\end{equation}
where $y$ is the target SELFIES token sequence, $\lambda$ weights the robustness term, and CE is cross-entropy loss. Equivalently, the token-level decoder loss can be written as
\begin{equation}
\mathcal{L}_{\mathrm{dec}}
=
-\sum_i
\left[
\log p_{\psi}(y_i \mid z_0, y_{<i})
+
\lambda \log p_{\psi}(y_i \mid \tilde{z}_0, y_{<i})
\right],
\end{equation}
where $i$ indexes the token positions in the target sequence, $y_i$ denotes the ground-truth SELFIES token at position $i$, and $y_{<i}$ denotes all preceding ground-truth tokens provided through teacher forcing. The distribution $p_{\psi}$ is the Transformer decoder's learned probability distribution over the SELFIES vocabulary. This objective encourages the decoder to reconstruct valid molecular strings not only from exact ChemBERTa encoder outputs but also from nearby latent points that may be produced during diffusion sampling.

\subsection{Latent diffusion model}

A denoising diffusion model is trained over the continuous ChemBERTa latent space. The forward process gradually perturbs a molecular embedding $z_0$ with Gaussian noise over $T$ timesteps:
\begin{equation}
q(z_t \mid z_0)
=
\mathcal{N}
\left(
z_t;
\sqrt{\bar{\alpha}_t} z_0,
(1-\bar{\alpha}_t)I
\right),
\end{equation}
where
\begin{equation}
\bar{\alpha}_t = \prod_{i=1}^{t}(1-\beta_i),
\end{equation}
and $\{\beta_t\}_{t=1}^{T}$ defines the noise schedule. In this work, a linear $\beta$ schedule is used with $T=500$ diffusion steps.

The denoising network $\epsilon_{\theta}$ is trained to predict the noise added to the latent embedding at each timestep. Given a sampled timestep $t$ and Gaussian noise $\epsilon \sim \mathcal{N}(0,I)$, the noised latent embedding is computed as
\begin{equation}
z_t =
\sqrt{\bar{\alpha}_t}z_0
+
\sqrt{1-\bar{\alpha}_t}\epsilon.
\end{equation}
The diffusion training objective is the standard denoising diffusion probabilistic model (DDPM) noise-prediction loss:
\begin{equation}
\mathcal{L}_{\mathrm{diff}}
=
\mathbb{E}_{z_0,t,\epsilon}
\left[
\left\|
\epsilon -
\epsilon_{\theta}(z_t,t)
\right\|_2^2
\right].
\end{equation}
After training, new molecular latent vectors are generated by sampling from a standard Gaussian prior and applying the learned reverse denoising process.

\subsection{Property prediction and guided sampling}

Property-guided generation is performed using differentiable predictors trained in the ChemBERTa latent space. For each target property, a binary property classifier $g_{\phi}$ estimates the probability that a molecule satisfies the desired endpoint. A separate latent guidance head $h_{\psi}$ is used during diffusion sampling to provide gradients for reverse-denoising guidance.

Given a ChemBERTa latent embedding $z$ and a binary property label $y \in \{0,1\}$, the property classifier predicts
\begin{equation}
\hat{y} = g_{\phi}(z),
\end{equation}
where $\hat{y} \in (0,1)$ is the predicted property probability. The latent guidance head estimates a differentiable property-guidance score, written as
\begin{equation}
\log p_{\psi}(y \mid z),
\end{equation}
whose gradient with respect to the latent variable is used during reverse diffusion.

The classifier and guidance head are trained using binary cross-entropy (BCE) objectives:
\begin{equation}
\mathcal{L}_{\mathrm{prop}}
=
\mathrm{BCE}\left(g_{\phi}(z), y\right)
+
\mathrm{BCE}\left(h_{\psi}(z), y\right).
\end{equation}
The property classifier $g_{\phi}$ is used for molecule scoring and reranking, while the latent guidance head $h_{\psi}$ is used to guide the reverse diffusion process.

During guided reverse denoising, the predicted reverse update is shifted in the direction that increases the target property likelihood. Following the notation in Figure~\ref{fig:overview}, the guided sampling step is written as
\begin{equation}
z_{t-1}
=
\mu_{\theta}(z_t,t)
+
\sigma_t^2
\nabla_z
\log p_{\psi}(y \mid z_t),
\end{equation}
where $\mu_{\theta}(z_t,t)$ is the reverse mean predicted by the diffusion model, $\sigma_t^2$ is the reverse-process variance at timestep $t$, and $\nabla_z \log p_{\psi}(y \mid z_t)$ is the latent-space guidance gradient provided by the guidance head. This update steers the reverse denoising trajectory toward regions of the ChemBERTa latent space that are associated with the desired molecular property.

\subsection{Molecular generation and evaluation}

Molecular generation begins by sampling latent vectors from a standard Gaussian distribution and applying the reverse diffusion process. The resulting latent vectors are decoded autoregressively into SELFIES strings using the Transformer decoder $D$. We use SELFIES as the decoder output representation because every valid SELFIES string maps to a chemically valid molecular graph, eliminating syntactic errors that commonly arise when generating SMILES directly and making the autoregressive decoding process more robust. The generated SELFIES strings are subsequently converted to SMILES for compatibility with standard cheminformatics tools and downstream evaluation.

Generated molecules are evaluated using standard molecular generation metrics. Validity measures the fraction of generated molecules that correspond to chemically valid RDKit molecules. Uniqueness measures the fraction of non-duplicate valid molecules among generated samples, and novelty measures the fraction of valid generated molecules that are not present in the training set. For guided generation, molecules are additionally scored using the trained binary property classifier $g_{\phi}$. We report the mean predicted property score, the success rate, defined as the fraction of valid generated molecules with predicted property score greater than 0.5, and the strict success rate, defined using a threshold of 0.8. We also report a property optimization score relative to a reference training sample to summarize the degree of property improvement induced by guidance.

To assess distributional similarity and diversity, descriptor-based metrics are computed using common RDKit molecular descriptors, including molecular weight, logP, topological polar surface area (TPSA), hydrogen-bond donors and acceptors, rotatable bonds, and quantitative estimate of drug-likeness (QED). Distributional differences between generated and reference molecules are summarized using mean descriptor KL divergence and descriptor-space maximum mean discrepancy. Molecular diversity is evaluated using internal diversity, while scaffold diversity is computed from Bemis--Murcko scaffolds \cite{bemis1996molecularframeworks} to characterize the structural diversity of generated molecules.

The evaluation metrics used throughout the Results section are summarized in Table~\ref{tab:metric-definitions}.

\begin{table}[t!]
    \centering
    \caption{Evaluation metrics for generated molecules. Arrows indicate the preferred direction.}
    \label{tab:metric-definitions}
    \small
    \renewcommand{\arraystretch}{1.25}
    \begin{tabularx}{\textwidth}{llcX}
        \toprule
        \textbf{Metric} & \textbf{Abbrev.} & \textbf{Direction} & \textbf{Definition} \\
        \midrule
        Validity & Validity & $\uparrow$ &
        Fraction of generated molecules that pass RDKit parsing, sanitization, and chemistry-consistency checks. \\

        Uniqueness & Unique & $\uparrow$ &
        Fraction of valid generated molecules that are unique after canonicalization. \\

        Novelty & Novel & $\uparrow$ &
        Fraction of valid generated molecules that are absent from the canonicalized training set. \\

        Mean property score & Mean score & $\uparrow$ &
        Average predicted target-property probability assigned by the trained classifier $g_{\phi}$ to valid generated molecules. \\

        Success rate & Success & $\uparrow$ &
        Fraction of valid generated molecules with predicted property score greater than 0.5. \\

        Strict success rate & Strict & $\uparrow$ &
        Fraction of valid generated molecules with predicted property score greater than 0.8. \\

        Property optimization & Opt. & $\uparrow$ &
        Difference between the mean generated property score and the mean reference-set property score. Positive values indicate improvement over the reference set. \\

        Internal diversity & Int. div. & $\uparrow$ &
        Mean pairwise molecular dissimilarity among valid generated molecules. \\

        Scaffold diversity & Scaf. div. & $\uparrow$ &
        Fraction of unique Bemis--Murcko scaffolds among valid generated molecules. \\

        Maximum mean discrepancy & MMD & $\downarrow$ &
        Descriptor-space discrepancy between generated and reference molecules computed from RDKit molecular descriptors. \\

        Mean Kullback–Leibler (KL) divergence & KL mean & $\downarrow$ &
        Mean Kullback–Leibler (KL) divergence between generated and reference descriptor distributions. \\
        \bottomrule
    \end{tabularx}
\end{table}



\section{Results}

We evaluated the proposed ChemBERTa latent diffusion framework on two property-guided molecular generation tasks: blood--brain barrier permeability (BBBP) and BACE-1 inhibition. A single decoder and latent diffusion model were trained using ZINC molecules and reused for both the BBBP and BACE-1 guidance experiments. Task-specific labeled datasets were used only to train the corresponding property classifiers and latent guidance heads. Generation was evaluated across six guidance strengths, $g \in \{0,0.5,1,2,3,5\}$, with 1,024 generated molecules per setting. All molecular-quality and property metrics were computed after generation using RDKit-based post-processing and the trained property classifier.

\subsection{Model Training}

Table~\ref{tab:training-metrics} summarizes the training performance of the shared decoder and latent diffusion model used in both property-guidance experiments. The decoder achieved a training token accuracy of 0.951 and a validation token accuracy of 0.935, indicating reliable ChemBERTa-to-SELFIES reconstruction. The latent diffusion model achieved a training noise-prediction loss of 0.0789 and a validation loss of 0.0797, indicating stable denoising performance. Because the same decoder and diffusion model are reused for both BBBP and BACE-1 guidance, their training results are reported only once.

\begin{table}[!t] 
    \centering 
    \caption{Training metrics for the shared decoder and latent diffusion model used in both BBBP and BACE-1 guidance experiments. Decoder performance is reported using token accuracy, while diffusion performance is reported using the DDPM noise-prediction loss.} 
    \label{tab:training-metrics} 
    \small 
        \begin{tabular}{lccccc} 
        \toprule 
        \textbf{Module} & \textbf{Epochs} & \textbf{Train loss} & \textbf{Val. loss} & \textbf{Train metric} & \textbf{Val. metric} \\ 
        \midrule 
        Decoder & 10 & 0.220 & 0.297 & token acc. 0.951 & token acc. 0.935 \\ 
        Diffusion & 10 & 0.0789 & 0.0797 & n/a & n/a \\ 
        \bottomrule 
        \end{tabular} 
\end{table}

Table~\ref{tab:property-predictor-comparison} provides a contextual comparison of the property classifiers used in our guidance framework with representative molecular property-prediction models. The proposed ChemBERTa-based classifier achieved validation ROC-AUC values of 0.904 for BBBP and 0.746 for BACE-1. The BBBP classifier shows strong predictive performance relative to the representative models included in the comparison, whereas BACE-1 remains a more challenging property-prediction task in our experimental setting.

\begin{table}[!t]
\centering
\caption{Contextual comparison of property-prediction ROC-AUC with representative molecular property predictors. Higher values are better.}
\label{tab:property-predictor-comparison}
\small
\begin{tabular}{lcc}
\toprule
\textbf{Model} & \textbf{BBBP AUC} & \textbf{BACE-1 AUC} \\
\midrule
ChemBERTa-2 \cite{ahmad2022chemberta2} & 0.701 & 0.809 \\
MolPLA \cite{gim2024molpla} & 0.702 & 0.805 \\
GROVER \cite{rong2020grover} & 0.936 & 0.878 \\
MolCLR \cite{wang2022molclr} & 0.736 & 0.890 \\
GraphMVP \cite{liu2022graphmvp} & 0.724 & 0.812 \\
MMSA \cite{yin2025mmsa} & 0.745 & 0.824 \\
LLaMA-2 (7B) \cite{touvron2023llama2} & 0.656 & 0.748 \\
LLM-MPP \cite{jin2025llmmpp} & 0.973 & 0.874 \\
\midrule
\textbf{Ours (ChemBERTa classifier)} & 0.904 & 0.746 \\
\bottomrule
\end{tabular}
\end{table}

\subsection{Generation Protocol}

A total of 1,024 molecules were generated across the four sampling seeds. The reported generation batches therefore contain samples generated across three independent sampling seeds. The trained model parameters and the train/validation data split were kept fixed across these generation runs; the different seeds affect the stochastic generation process only and do not correspond to independently retrained models. The generated molecules were subsequently validated, canonicalized, and scored in a separate evaluation stage.

\subsection{BBBP Guidance Results}

Table~\ref{tab:bbbp-results} reports the BBBP generation results. All selected molecules that entered the final evaluation passed RDKit validation and were novel with respect to the training set. Uniqueness was also consistently high, ranging from 0.998 to 1.000. Increasing guidance generally improved predicted BBBP activity. The mean classifier score increased from 0.811 at $g=0.0$ to 0.874 at $g=5.0$, while strict success increased from 0.668 to 0.800. The best overall property performance was obtained at $g=5.0$, which achieved the highest mean score, strict success rate, and optimization score. The highest standard success rate was shared by $g=3.0$ and $g=5.0$.

\begin{table}[H]
    \centering
    \caption{BBBP generation results across guidance strengths, obtained from generation using three random sampling seeds.}
    \label{tab:bbbp-results}
    \scriptsize
    \begin{tabularx}{\textwidth}{lXXXXXXXXX}
        \toprule
        \textbf{Guidance} &
        \textbf{Validity} &
        \textbf{Unique} &
        \textbf{Novel} &
        \textbf{Success} &
        \textbf{Strict} &
        \textbf{Mean score} &
        \textbf{Opt.} &
        \textbf{Int. div.} &
        \textbf{Scaf. div.} \\
        \midrule
        $g=0.0$ & 1.000 & 0.999 & 1.000 & 0.892 & 0.668 & 0.811 & 0.117 & 0.899 & \textbf{0.876} \\
        $g=0.5$ & 1.000 & 0.998 & 1.000 & 0.915 & 0.707 & 0.830 & 0.136 & 0.900 & 0.864 \\
        $g=1.0$ & 1.000 & 0.998 & 1.000 & 0.902 & 0.708 & 0.826 & 0.132 & \textbf{0.901} & 0.839 \\
        $g=2.0$ & 1.000 & \textbf{1.000} & 1.000 & 0.934 & 0.706 & 0.838 & 0.144 & 0.897 & 0.868 \\
        $g=3.0$ & 1.000 & \textbf{1.000} & 1.000 & \textbf{0.947} & 0.785 & 0.864 & 0.169 & 0.897 & 0.851 \\
        $g=5.0$ & 1.000 & 0.999 & 1.000 & \textbf{0.947} & \textbf{0.800} & \textbf{0.874} & \textbf{0.180} & 0.896 & 0.849 \\
        \bottomrule
    \end{tabularx}
\end{table}

\begin{figure}[H]
    \centering
    \includegraphics[width=0.90\textwidth]{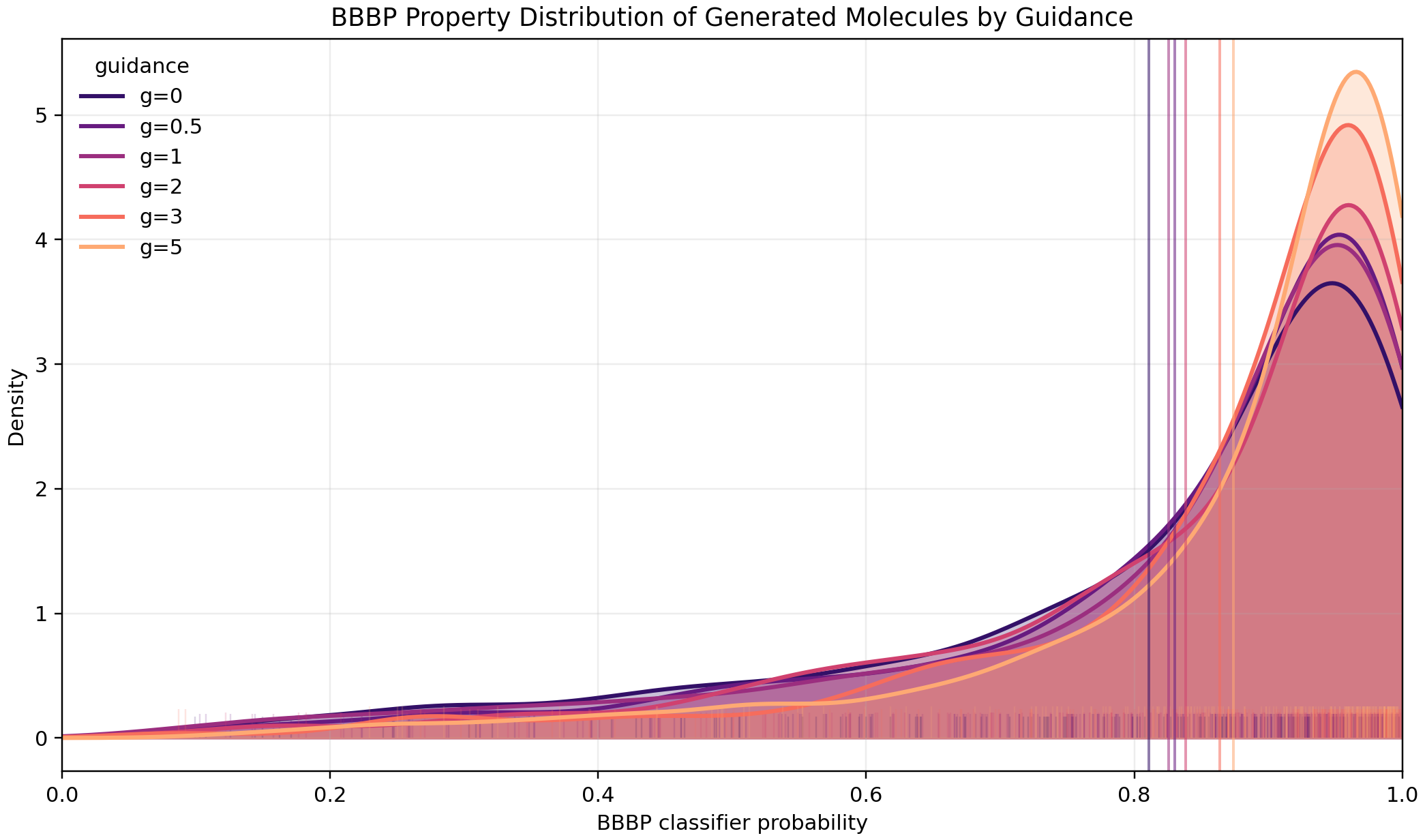}
    \caption{BBBP classifier-score distributions across guidance strengths. Higher guidance shifts generated molecules toward stronger predicted BBBP scores.}
    \label{fig:bbbp-property-distribution}
\end{figure}

Figure~\ref{fig:bbbp-property-distribution} and Table~\ref{tab:bbbp-results} show that BBBP guidance consistently shifts generated molecules toward higher predicted BBBP probabilities while maintaining high selected-molecule validity, uniqueness, and novelty. The strongest guidance setting, $g=5.0$, achieved the highest property-oriented performance, including the highest mean score, strict success rate, and optimization score. Scaffold diversity, however, was greatest without guidance ($g=0.0$), indicating that stronger property guidance can narrow the range of generated scaffolds. As shown in Figure~\ref{fig:trends}, guidance also changes the descriptor profile of generated molecules. For BBBP, increasing guidance generally lowers mean molecular weight and TPSA while increasing QED. Distribution-matching metrics favor intermediate guidance values, with the lowest MMD at $g=3.0$ and the lowest mean KL divergence at $g=2.0$. These results suggest that stronger BBBP guidance improves predicted property enrichment but also shifts the generated chemical distribution.

\subsection{BACE-1 Guidance Results}

Table~\ref{tab:bace-results} reports the BACE-1 generation results. As in the BBBP experiment, all selected molecules that entered the final evaluation passed RDKit validation and were novel with respect to the training set, and uniqueness remained between 0.999 and 1.000. Compared with BBBP, BACE-1 was a more difficult property target: baseline success and mean property scores were substantially lower. Nevertheless, strong guidance improved BACE-1-directed generation. The mean classifier score increased from 0.151 at $g=0.0$ to 0.213 at $g=5.0$, and the success rate increased from 0.089 to 0.138. The strongest guidance setting, $g=5.0$, achieved the best success rate, strict success rate, mean score, and optimization score.

\begin{table}[H]
    \centering
    \caption{BACE-1 generation results across guidance strengths, obtained from generation using three random sampling seeds.}
    \label{tab:bace-results}
    \scriptsize
    \begin{tabularx}{\textwidth}{lXXXXXXXXX}
        \toprule
        \textbf{Guidance} &
        \textbf{Validity} &
        \textbf{Unique} &
        \textbf{Novel} &
        \textbf{Success} &
        \textbf{Strict} &
        \textbf{Mean score} &
        \textbf{Opt.} &
        \textbf{Int. div.} &
        \textbf{Scaf. div.} \\
        \midrule
        $g=0.0$ & 1.000 & \textbf{1.000} & 1.000 & 0.089 & 0.014 & 0.151 & 0.036 & \textbf{0.899} & 0.867 \\
        $g=0.5$ & 1.000 & 0.999 & 1.000 & 0.084 & 0.011 & 0.157 & 0.041 & \textbf{0.899} & 0.855 \\
        $g=1.0$ & 1.000 & \textbf{1.000} & 1.000 & 0.110 & 0.018 & 0.170 & 0.054 & \textbf{0.899} & \textbf{0.886} \\
        $g=2.0$ & 1.000 & 0.999 & 1.000 & 0.097 & 0.013 & 0.170 & 0.054 & 0.897 & 0.873 \\
        $g=3.0$ & 1.000 & \textbf{1.000} & 1.000 & 0.105 & 0.018 & 0.190 & 0.074 & 0.896 & 0.866 \\
        $g=5.0$ & 1.000 & \textbf{1.000} & 1.000 & \textbf{0.138} & \textbf{0.023} & \textbf{0.213} & \textbf{0.097} & 0.897 & 0.872 \\
        \bottomrule
    \end{tabularx}
\end{table}

\begin{figure}[H]
    \centering
    \includegraphics[width=0.99\textwidth]{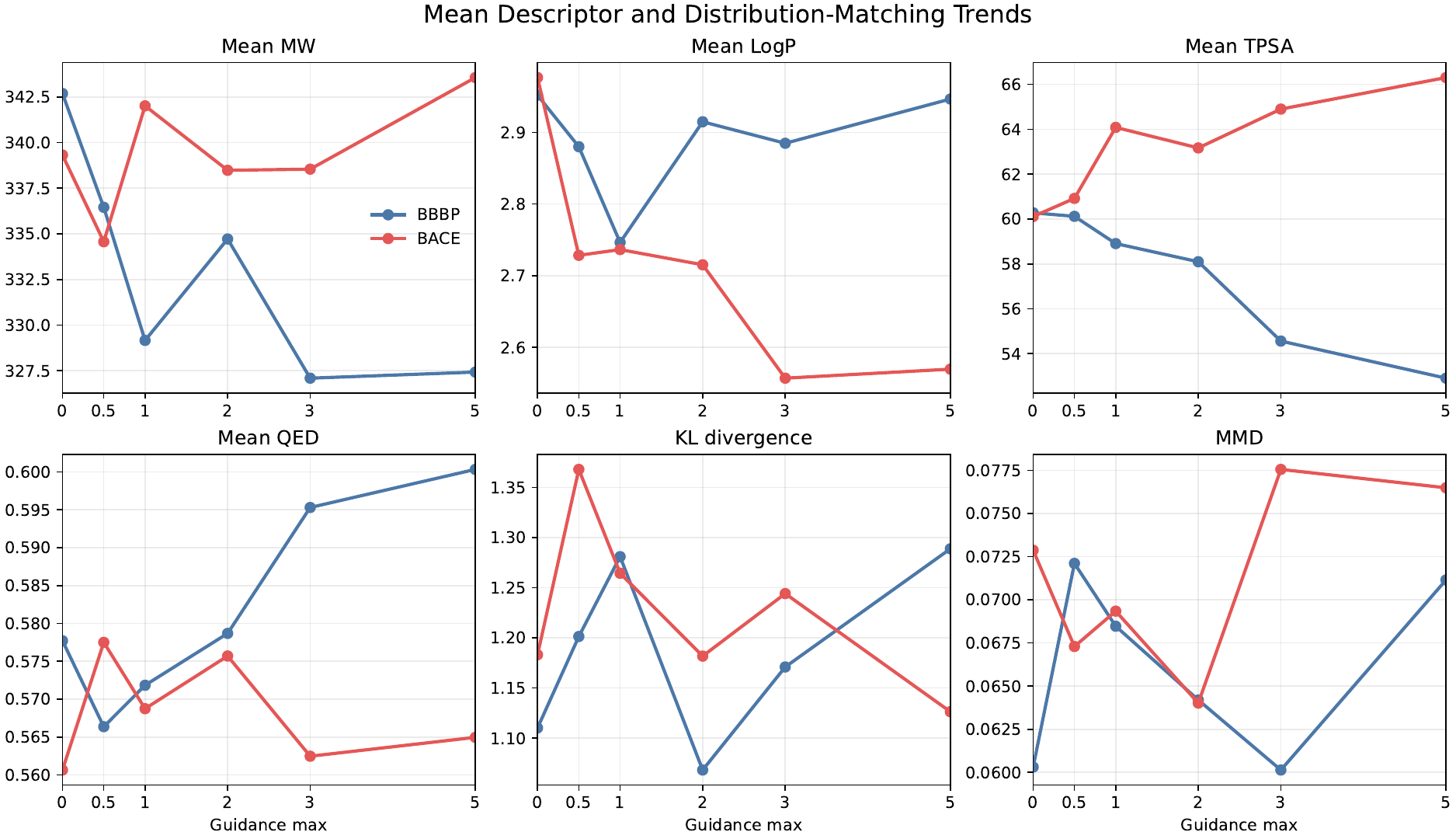}
    \caption{Mean descriptor and distribution-matching trends across guidance strengths for BBBP and BACE-1. The figure reports average molecular weight, logP, TPSA, and QED, together with descriptor-space KL divergence and maximum mean discrepancy (MMD). Lower KL divergence and MMD indicate closer agreement with the reference distribution.}
    \label{fig:trends}
\end{figure}

BACE-1 guidance produced smaller absolute enrichment than BBBP, but the strongest guidance setting still gave the best property-oriented metrics. Diversity was largely preserved during this shift: internal diversity remained between 0.896 and 0.899, and scaffold diversity remained between 0.855 and 0.886. The highest scaffold diversity occurred at $g=1.0$, whereas the strongest property metrics occurred at $g=5.0$. Figure~\ref{fig:trends} shows that BACE-1 guidance changes the descriptor profile differently from BBBP, with higher guidance tending to increase molecular weight and TPSA while reducing logP. Distribution-matching metrics also favored different operating points, with the lowest MMD at $g=2.0$ and the lowest mean KL divergence at $g=5.0$.

\subsection{Comparison Across Property Targets}

The two guidance tasks show different levels of difficulty. BBBP had high baseline enrichment, with an unguided success rate of 0.892 and mean classifier score of 0.811. In contrast, BACE-1 started from a substantially lower baseline, with an unguided success rate of 0.089 and mean classifier score of 0.151. Despite this difference, both tasks benefited from stronger guidance. At $g=5.0$, BBBP reached a mean score of 0.874 and strict success rate of 0.800, while BACE-1 reached a mean score of 0.213 and strict success rate of 0.023.

Across all settings, the selected molecules showed strong chemical-generation quality: evaluation-time validity and novelty were 1.000 for both tasks and all guidance values, and uniqueness never dropped below 0.998. These results indicate that ChemBERTa latent diffusion can preserve molecular validity and novelty while allowing task-specific guidance to shift the generated distribution toward desired classifier-defined properties. At the same time, the differences between property scores, scaffold diversity, and descriptor-distribution trends in Figure~\ref{fig:trends} suggest that guidance strength should be selected as a multi-objective hyperparameter rather than optimized solely for classifier score.

\section{Discussion}

The results show that ChemBERTa latent diffusion can generate high-quality selected molecules while enabling property-guided control. Across both BBBP and BACE-1, molecules that entered final evaluation achieved validity and novelty of 1.000, with uniqueness remaining at or above 0.998. This suggests that the combination of ChemBERTa embeddings and a SELFIES decoder provides a stable latent-to-molecule generation interface.

Latent guidance improved predicted property enrichment for both targets. For BBBP, the strongest guidance setting, $g=5.0$, produced the highest mean property score, strict success rate, and optimization score. BACE-1 was a more difficult endpoint, with lower absolute scores, but $g=5.0$ again gave the best property-oriented performance. These results indicate that a separate latent guidance head can steer generation toward target-specific properties without retraining the diffusion model.

The property-guided generation results are supported by the validation performance of the corresponding prediction and guidance models. For BBBP, the classifier and latent guidance head achieved strong scaffold-split validation ROC-AUC values of 0.904 and 0.899, respectively. The BACE-1 classifier and latent guidance head achieved moderate validation ROC-AUC values of 0.746 and 0.753, respectively. These results indicate stronger predictive generalization for the BBBP models while still providing validation support for the BACE-1 guidance experiment. Nevertheless, all reported property outcomes are model-predicted scores rather than experimental measurements and should therefore be interpreted accordingly.

Finally, stronger guidance was not uniformly optimal across all metrics. Although high guidance improved predicted property scores, scaffold diversity was better at lower or intermediate guidance settings, and Figure~\ref{fig:trends} shows that guidance also shifts molecular descriptors and distribution-matching metrics. Therefore, guidance strength should be selected as a multi-objective hyperparameter that balances predicted property enrichment, diversity, and distributional similarity. Future work should add stronger baselines, additional independent validation of the property predictors and latent guidance heads, and raw candidate validity analysis.

\section{Conclusions}

This work presented a ChemBERTa latent diffusion framework for property-guided small-molecule generation. The proposed pipeline combines frozen ChemBERTa embeddings, a Transformer-based SELFIES decoder, an unconditional latent diffusion model, and task-specific latent guidance heads for steering generation toward desired molecular properties. Across BBBP and BACE-1 guidance experiments, the selected generated molecules maintained high chemical quality, with validity and novelty of 1.000 and uniqueness of at least 0.998 across all guidance settings.

The guidance results show that latent-space steering can improve predicted property enrichment without retraining the diffusion model. Strong guidance produced the best property-oriented metrics for both targets, with $g=5.0$ achieving the highest mean score, strict success rate, and optimization score in the reported sweeps. However, the results also show that guidance strength affects diversity and descriptor-distribution behavior. Scaffold diversity and distribution-matching metrics did not always favor the same guidance setting as the property scores, indicating that guidance should be selected as a multi-objective hyperparameter rather than optimized only for classifier confidence.

Overall, the study supports the feasibility of using pretrained molecular language-model embeddings as a continuous latent space for controllable molecular generation. Future work should strengthen the evaluation with independently validated property predictors, raw candidate validity analysis, repeated sampling runs, and comparisons against graph-based, language-model, and VAE-based molecular generation baselines.

\vspace{1.5em}
{
\small
\setlength{\parskip}{0.8em}

\noindent \textbf{Supplementary Materials:} The project repository is publicly available at github: \\ \href{https://github.com/FlockOfBird/GMGD}{https://github.com/FlockOfBird/GMGD}.

\noindent \textbf{Author Contributions:} Conceptualization, A.F.; methodology, A.F., L.R., A.N.; validation, A.F.; formal analysis, A.F., L.R., and A.N.; writing--original draft preparation, A.F.; writing--review and editing, A.F., L.R., and A.N.; supervision. L.R., A.N.; project management, L.R., A.N.;

\noindent \textbf{Funding:} This research has been partially funded by the Natural Sciences and Engineering Research Council of Canada, NSERC.

\noindent \textbf{Data Availability Statement:} The experiments use publicly available ZINC, BACE-1 and BBBP datasets.

\noindent \textbf{Conflicts of Interest:} The author declares no conflict of interest.
\par
}

\bibliographystyle{ieeetr}
\bibliography{references}

@ARTICLE{wang2025diffmolsurvey,
    author = {{Wang}, Liang and {Song}, Chao and {Liu}, Zhiyuan and {Rong}, Yu and {Liu}, Qiang and {Wu}, Shu and {Wang}, Liang},
    title = "{Diffusion Models for Molecules: A Survey of Methods and Tasks}",
    journal = {arXiv e-prints},
    year = 2025,
    month = feb,
    eid = {arXiv:2502.09511},
    pages = {arXiv:2502.09511},
    doi = {10.48550/arXiv.2502.09511},
    archivePrefix = {arXiv},
    eprint = {2502.09511},
    primaryClass = {cs.LG},
    adsurl = {https://ui.adsabs.harvard.edu/abs/2025arXiv250209511W}
}

@article{Qin20253dsurvey,
    author = {Qin, Yifei and Wei, Xuexin and Xu, Mingyuan and Wu, Jiaqiang and Tang, Miru and Ran, Ting and Chen, Hongming},
    title = {Comprehensive Benchmark Study of Diffusion-Based 3D Molecular Generation Models},
    journal = {ACS Omega},
    volume = {10},
    number = {37},
    pages = {42760-42775},
    year = {2025},
    doi = {10.1021/acsomega.5c05077},
    URL = {https://doi.org/10.1021/acsomega.5c05077},
    eprint = {https://doi.org/10.1021/acsomega.5c05077}
}

@article{alakhdar2024diffsurvey,
    author = {Alakhdar, Amira and Poczos, Barnabas and Washburn, Newell},
    title = {Diffusion Models in De Novo Drug Design},
    journal = {Journal of Chemical Information and Modeling},
    volume = {64},
    number = {19},
    pages = {7238-7256},
    year = {2024},
    doi = {10.1021/acs.jcim.4c01107},
    note ={PMID: 39322943},
    URL = {https://doi.org/10.1021/acs.jcim.4c01107},
    eprint = {https://doi.org/10.1021/acs.jcim.4c01107}
}

@article{benjamin2018inversesurvey,
    author = {Benjamin Sanchez-Lengeling  and Alán Aspuru-Guzik },
    title = {Inverse molecular design using machine learning: Generative models for matter engineering},
    journal = {Science},
    volume = {361},
    number = {6400},
    pages = {360-365},
    year = {2018},
    doi = {10.1126/science.aat2663},
    URL = {https://www.science.org/doi/abs/10.1126/science.aat2663},
    eprint = {https://www.science.org/doi/pdf/10.1126/science.aat2663},
}

@article{wu2018moleculenet,
  title={MoleculeNet: a benchmark for molecular machine learning},
  author={Wu, Zhenqin and Ramsundar, Bharath and Feinberg, Evan N. and Gomes, Joseph and Geniesse, Caleb and Pappu, Aneesh S. and Leswing, Karl and Pande, Vijay},
  journal={Chemical Science},
  volume={9},
  number={2},
  pages={513--530},
  year={2018},
  publisher={Royal Society of Chemistry},
  doi={10.1039/C7SC02664A}
}

@article{martins2012bbbp,
    author = {Martins, Ines Filipa and Teixeira, Ana L. and Pinheiro, Luis and Falcao, Andre O.},
    title = {{A Bayesian Approach to in Silico Blood-Brain Barrier Penetration Modeling}},
    journal = {Journal of Chemical Information and Modeling},
    volume = {52},
    number = {6},
    pages = {1686-1697},
    year = {2012},
    doi = {10.1021/ci300124c},
    note ={PMID: 22612593},
    URL = {https://doi.org/10.1021/ci300124c},
}

@article{subramanian2016bace,
    author = {Subramanian, Govindan and Ramsundar, Bharath and Pande, Vijay and Denny, Rajiah Aldrin},
    title = {{Computational Modeling of {$\beta$}-Secretase 1 (BACE-1) Inhibitors Using Ligand Based Approaches}},
    journal = {Journal of Chemical Information and Modeling},
    volume = {56},
    number = {10},
    pages = {1936-1949},
    year = {2016},
    doi = {10.1021/acs.jcim.6b00290},
    note = {PMID: 27689393},
    URL = {https://doi.org/10.1021/acs.jcim.6b00290},
    eprint = {https://doi.org/10.1021/acs.jcim.6b00290}
}

@article{irwin2020zinc,
    title = {{ZINC20—A Free Ultralarge-Scale Chemical Database for Ligand Discovery}},
    author = {Irwin, John J. and Tang, Khanh G. and Young, Jennifer and Dandarchuluun, Chinzorig and Wong, Benjamin R. and Khurelbaatar, Munkhzul and Moroz, Yurii S. and Mayfield, John and Sayle, Roger A.},
    journal = {Journal of Chemical Information and Modeling},
    volume = {60},
    number = {12},
    pages = {6065-6073},
    year = {2020},
    doi = {10.1021/acs.jcim.0c00675},
    note ={PMID: 33118813},
    url = {https://doi.org/10.1021/acs.jcim.0c00675},
    eprint = { https://doi.org/10.1021/acs.jcim.0c00675}
}

@article{weininger1988smiles,
  title={SMILES, a chemical language and information system. 1. Introduction to methodology and encoding rules},
  author={Weininger, David},
  journal={Journal of Chemical Information and Computer Sciences},
  volume={28},
  number={1},
  pages={31--36},
  year={1988},
  doi={10.1021/ci00057a005}
}

@article{weininger1989smiles,
  title={SMILES. 2. Algorithm for generation of unique SMILES notation},
  author={Weininger, David and Weininger, Arthur and Weininger, Joseph L.},
  journal={Journal of Chemical Information and Computer Sciences},
  volume={29},
  number={2},
  pages={97--101},
  year={1989},
  doi={10.1021/ci00062a008}
}

@article{krenn2020selfies,
  title={Self-referencing embedded strings (SELFIES): A 100\% robust molecular string representation},
  author={Krenn, Mario and H{\"a}se, Florian and Nigam, AkshatKumar and Friederich, Pascal and Aspuru-Guzik, Al{\'a}n},
  journal={Machine Learning: Science and Technology},
  volume={1},
  number={4},
  pages={045024},
  year={2020},
  doi={10.1088/2632-2153/aba947}
}

@article{chithrananda2020chemberta,
  title={ChemBERTa: Large-Scale Self-Supervised Pretraining for Molecular Property Prediction},
  author={Chithrananda, Seyone and Grand, Gabriel and Ramsundar, Bharath},
  journal={arXiv preprint arXiv:2010.09885},
  year={2020}
}

@article{ahmad2022chemberta2,
  title={ChemBERTa-2: Towards Chemical Foundation Models},
  author={Ahmad, Walid and Simon, Elana and Chithrananda, Seyone and Grand, Gabriel and Ramsundar, Bharath},
  journal={arXiv preprint arXiv:2209.01712},
  year={2022}
}

@article{Du2024MachineLG,
    title={Machine learning-aided generative molecular design},
    author={Yuanqi Du and Arian R. Jamasb and Jeff Guo and Tianfan Fu and Charles Harris and Yingheng Wang and Chenru Duan and Pietro Li{\'o} and Philippe Schwaller and Tom L. Blundell},
    journal={Nature Machine Intelligence},
    year={2024},
    volume={6},
    pages={589 - 604},
    url={https://api.semanticscholar.org/CorpusID:270606382}
}

@article{SohlDickstein2015DeepUL,
    title={Deep Unsupervised Learning using Nonequilibrium Thermodynamics},
    author={Jascha Narain Sohl-Dickstein and Eric A. Weiss and Niru Maheswaranathan and Surya Ganguli},
    journal={ArXiv},
    year={2015},
    volume={abs/1503.03585},
    url={https://api.semanticscholar.org/CorpusID:14888175}
}

@inproceedings{Ho2020denoisingdiff,
    author = {Ho, Jonathan and Jain, Ajay and Abbeel, Pieter},
    title = {Denoising diffusion probabilistic models},
    year = {2020},
    isbn = {9781713829546},
    publisher = {Curran Associates Inc.},
    address = {Red Hook, NY, USA},
    booktitle = {Proceedings of the 34th International Conference on Neural Information Processing Systems},
    articleno = {574},
    numpages = {12},
    location = {Vancouver, BC, Canada},
    series = {NIPS '20}
}

@inproceedings{rombach2022highresimgae,
    author = { Rombach, Robin and Blattmann, Andreas and Lorenz, Dominik and Esser, Patrick and Ommer, Bjorn },
    booktitle = { 2022 IEEE/CVF Conference on Computer Vision and Pattern Recognition (CVPR) },
    title = {{ High-Resolution Image Synthesis with Latent Diffusion Models }},
    year = {2022},
    volume = {},
    ISSN = {},
    pages = {10674-10685},
    doi = {10.1109/CVPR52688.2022.01042},
    url = {https://doi.ieeecomputersociety.org/10.1109/CVPR52688.2022.01042},
    publisher = {IEEE Computer Society},
    address = {Los Alamitos, CA, USA},
    month =Jun
}

@inproceedings{Vignac2023MiDiMG,
    title={MiDi: Mixed Graph and 3D Denoising Diffusion for Molecule Generation},
    author={Cl{\'e}ment Vignac and Nagham Osman and Laura Davison Mangilli Toni and Pascal Frossard},
    booktitle={ECML/PKDD},
    year={2023},
    url={https://api.semanticscholar.org/CorpusID:257020078}
}

@article{Liu2025NExTMol3D,
    title={NExT-Mol: 3D Diffusion Meets 1D Language Modeling for 3D Molecule Generation},
    author={Zhiyuan Liu and Yanchen Luo and Han Huang and Enzhi Zhang and Sihang Li and Junfeng Fang and Yaorui Shi and Xiang Wang and Kenji Kawaguchi and Tat-Seng Chua},
    journal={ArXiv},
    year={2025},
    volume={abs/2502.12638},
    url={https://api.semanticscholar.org/CorpusID:276422386}
}

@article{Feng2024UniGem,
    author = {{Feng}, Shikun and {Ni}, Yuyan and {Lu}, Yan and {Ma}, Zhi-Ming and {Ma}, Wei-Ying and {Lan}, Yanyan},
    title = "{UniGEM: A Unified Approach to Generation and Property Prediction for Molecules}",
    journal = {arXiv e-prints},
    year = 2024,
    month = oct,
    eid = {arXiv:2410.10516},
    pages = {arXiv:2410.10516},
    doi = {10.48550/arXiv.2410.10516},
    archivePrefix = {arXiv},
    eprint = {2410.10516},
    primaryClass = {stat.ML},
    adsurl = {https://ui.adsabs.harvard.edu/abs/2024arXiv241010516F}
}

@article{Lee2022ExploringCS,
    title={Exploring Chemical Space with Score-based Out-of-distribution Generation},
    author={Seul Lee and Jaehyeong Jo and Sung Ju Hwang},
    journal={ArXiv},
    year={2022},
    volume={abs/2206.07632},
    url={https://api.semanticscholar.org/CorpusID:249674702}
}

@article{Qiang2023CoarsetoFineAH,
  title={Coarse-to-Fine: a Hierarchical Diffusion Model for Molecule Generation in 3D},
  author={Bo Qiang and Yuxuan Song and Minkai Xu and Jingjing Gong and Bo Gao and Hao Zhou and Weiying Ma and Yanyan Lan},
  journal={ArXiv},
  year={2023},
  volume={abs/2305.13266},
  url={https://api.semanticscholar.org/CorpusID:258833390}
}

@inproceedings{liu2024graphdit,
    title={Graph Diffusion Transformers for Multi-Conditional Molecular Generation},
    author={Liu, Gang and Xu, Jiaxin and Luo, Tengfei and Jiang, Meng},
    booktitle={The Thirty-eighth Annual Conference on Neural Information Processing Systems},
    year={2024}
}

@article{Hua2023MUDiffUD,
    title={MUDiff: Unified Diffusion for Complete Molecule Generation},
    author={Chenqing Hua and Sitao Luan and Minkai Xu and Rex Ying and Jie Fu and Stefano Ermon and Doina Precup},
    journal={ArXiv},
    year={2023},
    volume={abs/2304.14621},
    url={https://api.semanticscholar.org/CorpusID:258417871}
}

@inproceedings{huang2025midi,
    title={Midi: Multi-instance diffusion for single image to 3d scene generation},
    author={Huang, Zehuan and Guo, Yuan-Chen and An, Xingqiao and Yang, Yunhan and Li, Yangguang and Zou, Zi-Xin and Liang, Ding and Liu, Xihui and Cao, Yan-Pei and Sheng, Lu},
    booktitle={Proceedings of the Computer Vision and Pattern Recognition Conference},
    pages={23646--23657},
    year={2025}
}

@article{Peng2023MolDiffAT,
    title={MolDiff: Addressing the Atom-Bond Inconsistency Problem in 3D Molecule Diffusion Generation},
    author={Xingang Peng and Jiaqi Guan and Qiang Liu and Jianzhu Ma},
    journal={ArXiv},
    year={2023},
    volume={abs/2305.07508},
    url={https://api.semanticscholar.org/CorpusID:258676222}
}

@article{bemis1996molecularframeworks,
    title={The Properties of Known Drugs. 1. Molecular Frameworks},
    author={Bemis, Guy W. and Murcko, Mark A.},
    journal={Journal of Medicinal Chemistry},
    volume={39},
    number={15},
    pages={2887--2893},
    year={1996},
    doi={10.1021/jm9602928}
}

@article{Oestreich2024DrugDiffS,
    title={DrugDiff - small molecule diffusion model with flexible guidance towards molecular properties},
    author={Marie Oestreich and Erinc Merdivan and Michael Lee and Joachim L. Schultze and Marie Piraud and Matthias Becker},
    journal={bioRxiv},
    year={2024},
    url={https://api.semanticscholar.org/CorpusID:271325009}
}

@article{gim2024molpla,
  title={MolPLA: A Molecular Pretraining Framework for Learning Cores, R-Groups and Their Linker Joints},
  author={Gim, Mogan and Park, Jueon and Park, Soyon and Lee, Sanghoon and Baek, Seungheun and Lee, Junhyun and Nguyen, Ngoc-Quang and Kang, Jaewoo},
  journal={Bioinformatics},
  volume={40},
  number={Supplement\_1},
  pages={i369--i380},
  year={2024},
  doi={10.1093/bioinformatics/btae256}
}

@inproceedings{rong2020grover,
  title={Self-Supervised Graph Transformer on Large-Scale Molecular Data},
  author={Rong, Yu and Bian, Yatao and Xu, Tingyang and Xie, Weiyang and Wei, Ying and Huang, Wenbing and Huang, Junzhou},
  booktitle={Advances in Neural Information Processing Systems},
  volume={33},
  pages={12559--12571},
  year={2020}
}

@article{wang2022molclr,
  title={Molecular Contrastive Learning of Representations via Graph Neural Networks},
  author={Wang, Yuyang and Wang, Jianren and Cao, Zhonglin and Barati Farimani, Amir},
  journal={Nature Machine Intelligence},
  volume={4},
  pages={279--287},
  year={2022},
  doi={10.1038/s42256-022-00447-x}
}

@inproceedings{liu2022graphmvp,
  title={Pre-Training Molecular Graph Representation with 3D Geometry},
  author={Liu, Shengchao and Wang, Hanchen and Liu, Weiyang and Lasenby, Joan and Guo, Hongyu and Tang, Jian},
  booktitle={International Conference on Learning Representations},
  year={2022},
  url={https://openreview.net/forum?id=xQUe1pOKPam}
}

@article{yin2025mmsa,
  title={Multi-Modal Molecular Representation Learning via Structure Awareness},
  author={Yin, Rong and Liu, Ruyue and Hao, Xiaoshuai and Zhou, Xingrui and Liu, Yong and Ma, Can and Wang, Weiping},
  journal={IEEE Transactions on Image Processing},
  volume={34},
  pages={3225--3238},
  year={2025},
  doi={10.1109/TIP.2025.3570604}
}

@article{touvron2023llama2,
  title={Llama 2: Open Foundation and Fine-Tuned Chat Models},
  author={Touvron, Hugo and Martin, Louis and Stone, Kevin and Albert, Peter and Almahairi, Amjad and Babaei, Yasmine and Bashlykov, Nikolay and Batra, Soumya and Bhargava, Prajjwal and Bhosale, Shruti and Bikel, Dan and Blecher, Lukas and Ferrer, Cristian Canton and Chen, Moya and Cucurull, Guillem and Esiobu, David and Fernandes, Jude and Fu, Jeremy and Fu, Wenyin and Fuller, Brian and Gao, Cynthia and Goswami, Vedanuj and Goyal, Naman and Hartshorn, Anthony and Hosseini, Saghar and Hou, Rui and Inan, Hakan and Kardas, Marcin and Kerkez, Viktor and Khabsa, Madian and Kloumann, Isabel and Korenev, Artem and Koura, Punit Singh and Lachaux, Marie-Anne and Lavril, Thibaut and Lee, Jenya and Liskovich, Diana and Lu, Yinghai and Mao, Yuning and Martinet, Xavier and Mihaylov, Todor and Mishra, Pushkar and Molybog, Igor and Nie, Yixin and Poulton, Andrew and Reizenstein, Jeremy and Rungta, Rashi and Saladi, Kalyan and Schelten, Alan and Silva, Ruan and Smith, Eric Michael and Subramanian, Ranjan and Tan, Xiaoqing Ellen and Tang, Binh and Taylor, Ross and Williams, Adina and Kuan, Jian Xiang and Xu, Puxin and Yan, Zheng and Zarov, Iliyan and Zhang, Yuchen and Fan, Angela and Kambadur, Melanie and Narang, Sharan and Rodriguez, Aurelien and Stojnic, Robert and Edunov, Sergey and Scialom, Thomas},
  journal={arXiv preprint arXiv:2307.09288},
  year={2023}
}

@article{jin2025llmmpp,
  title={Effective and Explainable Molecular Property Prediction by Chain-of-Thought Enabled Large Language Models and Multi-Modal Molecular Information Fusion},
  author={Jin, Chang and Guo, Siyuan and Zhou, Shuigeng and Guan, Jihong},
  journal={Journal of Chemical Information and Modeling},
  volume={65},
  number={11},
  pages={5438--5455},
  year={2025},
  doi={10.1021/acs.jcim.5c00577}
}

\end{document}